\documentclass{article}

 \PassOptionsToPackage{square,numbers}{natbib}

\usepackage[final]{nips_2018}

\usepackage[utf8]{inputenc} 
\usepackage[T1]{fontenc}    
\usepackage{hyperref}       
\usepackage{url}            
\usepackage{booktabs}       
\usepackage{amsfonts}       
\usepackage{nicefrac}       
\usepackage{microtype}      
\usepackage{amsmath}
\usepackage{listings}
\usepackage{graphicx}
\graphicspath{ {./images/} }
\usepackage{subfig}

\title{Sensitivity based Neural Networks Explanations}

\author{
  Enguerrand Horel \\
  Institute for Computational and Mathematical Engineering\\
  Stanford University\\
  Stanford, CA 94230\\
  \texttt{ehorel@stanford.edu} \\
  \And 
  Virgile Mison \thanks{Opinions expressed in this paper are those of the authors, and do not necessarily reflect the view of J.P. Morgan.} \\
  Machine Learning Center of Excellence\\
  J.P. Morgan\\
  New York, NY 10017\\
    \texttt{virgile.ca.mison@jpmchase.com}\\
  \And 
  Tao Xiong \footnotemark[1]  \\
Machine Learning Center of Excellence\\
  J.P. Morgan\\
  New York, NY 10017\\
  \texttt{tao.xiong@jpmchase.com}\\
  \And
  Kay Giesecke\\
  Department of Management Science \& Engineering\\
  Stanford University\\
  Stanford, CA 94305\\
  \texttt{giesecke@stanford.edu} \\
    \And 
  Lidia Mangu \footnotemark[1] \\
Machine Learning Center of Excellence\\
  J.P. Morgan\\
  New York, NY 10017\\
  \texttt{lidia.l.mangu@jpmchase.com}\\
}

\begin{document}

\maketitle

\begin{abstract}
Although neural networks can achieve very high predictive performance on various different tasks such as image recognition or natural language processing, they are often considered as opaque "black boxes". The difficulty of interpreting the predictions of a neural network often prevents its use in fields where explainability is important, such as the financial industry where regulators and auditors often insist on this aspect. In this paper, we present a way to assess the relative input features importance of a neural network based on the sensitivity of the model output with respect to its input. This method has the advantage of being fast to compute, it can provide both global and local levels of explanations and is applicable for many types of neural network architectures. We illustrate the performance of this method on both synthetic and real data and compare it with other interpretation techniques. This method is implemented into an open-source Python package that allows its users to easily generate and visualize explanations for their neural networks.
\end{abstract}

\section{Introduction}

With the massive development of computing resources in the past twenty years, artificial neural networks have come back to the forefront of machine learning.  This new computing capacity facilitating training deep and complex networks on large datasets has led to the resurgence of Deep Learning. The current popularity of deep learning is mainly due to the outstanding predictive performance of neural networks in a lot of different applications such as image recognition \cite{krizhevsky2012imagenet}, natural language processing \cite{collobert2011natural}, speech recognition \cite{hinton2012deep} or genomics \cite{xiong2015human}. In finance, deep neural networks have been applied with success to assets pricing \cite{chen2018deep}, \cite{gu2018empirical}, limit order books \cite{sirignano2016deep}, portfolio selection \cite{heaton2016deep}, option pricing \cite{hutchinson1994nonparametric}, hedging \cite{buehler2018deep} and mortgage risk \cite{sirignano2016deepmortgage}. 

Although deep neural networks are better than most of the more traditional machine learning techniques for complex learning tasks, their lack of interpretability remains a major obstacle to their use in some fields. In medical applications for instance, the risk of using the outcome of a black-box model to perform a diagnostic is too high since a misdiagnosis could potentially have dramatic consequences \cite{eftekhar2005comparison}, \cite{caruana2015intelligible}. In credit scoring, the Equal Credit Opportunity Act requires lenders to be able to explain why they decide to reject a credit application and prove that they do not rely on discriminatory features \cite{trinkle2007interpretable}. Neural networks interpretability is indeed necessary for two main reasons: trust and informativeness. Interpretation that leads to a clear understanding of how a model generates its outputs is key for trusting it and make sure it will generalize well on new data. Besides, being able to interpret a model could potentially unravel properties of the data such as discovering causalities across variables. 

There are many types of explanations one could provide to interpret a model, in this paper we choose to focus on explaining the relative importance and significance of the input features of a model. We believe that this represents a very fundamental and probably the most common type of explanations one would expect to receive to better understand a model. As reviewed in \cite{steppe1997feature}, \cite{zhang2000neural}, several heuristic measures have been proposed to estimate the relative importance of input features of neural networks. Most of them can be classified into either derivative-based methods, i.e. methods measuring the relative changes in neural network output with respect to its inputs or weight-based methods that assess the value of a certain function of the weights corresponding to each feature. In \cite{lime}, a Local Interpretable Model-Agnostic Explanations (LIME) method is proposed to explain the predictions of any classifier. This method gives local types of explanations by generating a sparse linear model around a specific prediction. Another feature importance metric is proposed in \cite{lipovetsky2001analysis}, where the notion of Shapley Value imputation from co-operative game theory is used to define the Shapley regression values. This metric has the advantage of being particularly robust in the presence of multicollinearity among inputs but cannot be computed efficiently for high dimensional problems.

In this paper, we choose a feature importance metric based on the sensitivity of a neural network output with respect to its input. A statistical justification of this choice of metric and how it can be used to perform rigorous significance tests for neural network can be found in \cite{horel2018explainable}. While the idea of using the sensitivity of a model is not new and has been previously used for neural networks, we think that the contribution of this paper is to justify its use over other feature importance methods for the specific case of neural networks. We show how this metric, that is easy and fast to compute, can be used to provide both global and local levels of explanations and how it can be adapted to fit most of the existing types of neural network architectures such as fully-connected, convolutional or recurrent neural networks. The method described in this paper is implemented into an open-source Python package that allows its users to easily generate and visualize explanations for their neural networks.

In Section \ref{S::2}, we describe the various types of neural network architectures considered and their corresponding feature importance metrics. We illustrate the use of this method in a simulation setting in Section \ref{S::3}. Finally, we apply this method to the real case of customers credit card default payments in Section \ref{S::4}.

\section{Method}
\label{S::2}

\subsection{Overall interpretability strategy}
\label{SS::2.1}

In order to understand how a neural network model is making its predictions, we propose a sensitivity analysis based on the derivatives of the model outputs with respect to its inputs. We believe that a sensitivity based method is particularly well suited for explaining neural networks due to the following reasons.
\begin{itemize}
\item The sensitivity of a model output with respect to its inputs is a very intuitive way to describe a model.
\item Neural networks are trained using variants of stochastic gradient descent optimization techniques that require computation of the derivatives of the loss function with respect to the weights of the network. This means that neural networks are inherently differential (or at least sub-differentiable for networks based on some sub-differentiable activation functions such as ReLU) and that their derivatives can be easily computed by leveraging the tensor libraries that are used to train them. 
\item The derivative of a model with respect to its input is a very granular type of information that can be aggregated in many different ways, it can yield to a level of interpretation that can be local or global. For instance, by averaging the square of the derivative over all the training samples that have been used to fit the model, one can get information of what has been learned overall by the model. In contrast, computing the sensitivities at one particular observation informs on the model behavior locally around it. This local level of information is particularly useful for understanding the reasons for producing a certain output from a specific observation. 
\end{itemize}

\subsection{Types of Neural Networks Architectures}
\label{SS::2.2}

This sensitivity based method can be applied and adapted to different types of neural networks architectures. In each case, we assume that the neural network model $f$ maps input features $x$ to an outcome $y$ that can be a real number in case of a regression or a categorical variable for a classification task. We consider the following types of networks. 
\begin{itemize}
    \item Fully-connected or convolutional neural networks, that is architectures with no sequential nor temporal components.
    $$y^i = f(x^i) $$
    In this case, the input features $x^i$ of the $i$-th sample is either a vector as it is usually the case for fully-connected models but can also be a matrix or a 3-D tensor for convolutional neural networks.
    \item Many-to-one recurrent neural networks, that is architecture with a sequential dependency of the input features. Given a sequence length $\tau$ and an initial hidden state $h_0$, these models are defined in a recursive way as follow:
\begin{align*}
    y^t &= h^t_{\tau} = f(x^t_1,x^t_2,...,x^t_p,h^t_{\tau-1}) \\
    h^t_{\tau-1} &= f(x^{t-1}_1,x^{t-1}_2,...,x^{t-1}_p,h^t_{\tau-2}) \\
    &\vdots\\
    h^t_{1} &= f(x^{t-\tau+1}_1,x^{t-\tau+1}_2,...,x^{t-\tau+1}_p,h_0)
\end{align*}
In this case, the input feature space is 2 dimensional. The input features $x^t$ of the $t$-th sequence is formed of $p$ different input sequences each of length $\tau$.
\item Many-to-many recurrent neural networks, that is architecture with a sequential dependency in both input and outputs. 
\begin{align*}
    y^t &= h^t_{\tau} = f(x^t_1,x^t_2,...,x^t_p,h^t_{\tau-1}) \\
    y^{t-1} = h^t_{\tau-1} &= f(x^{t-1}_1,x^{t-1}_2,...,x^{t-1}_p,h^t_{\tau-2}) \\
    &\vdots\\
    y^{t-\tau+1} = h^t_{1} &= f(x^{t-\tau+1}_1,x^{t-\tau+1}_2,...,x^{t-\tau+1}_p,h_0)
\end{align*}
\end{itemize}

\subsection{Global feature importance}
\label{SS::2.3}

We are first presenting a way to measure the relative importance of input features at a global level. This allows the user to understand what has been learned by the neural network learns during training.
The global importance of input feature or sequence $j$  over the training dataset of the model is defined as follow depending on the type of architecture considered (as described in \ref{SS::2.2}).

\begin{equation*}
    \lambda_j = \frac{100}{C}\sqrt{\frac{1}{n} \sum_{i=1}^n \Bigg(\frac{\partial f(x^i)}{\partial x_j}\Bigg)^2}
\end{equation*}
\begin{equation*}
    \lambda_j = \frac{100}{C}\sqrt{\frac{1}{T} \sum_{t=1}^T \Bigg(\frac{\partial y^t}{\partial x_j^t}\Bigg)^2}
\end{equation*}
\begin{equation*}
    \lambda_j = \frac{100}{C\tau}\sum_{l=0}^{\tau-1} \sqrt{\frac{1}{T} \sum_{t=1}^T \Bigg(\frac{\partial y^{t-l}}{\partial x_j^{t-l}}\Bigg)^2}
\end{equation*}

$n$ is the number of training samples in the case of independent observations and $T$ represents the number of training sequences in the case of dependent observations. $C$ is the normalization factor so that $\sum_{j=1}^p \lambda_j = 100$, where $p$ is the number of input features. The derivatives are averaged across all the samples of the dataset to capture the global sensitivity and are squared to avoid cancellation of positive with negative values.

The main use of this metric is to rank the features by predictive power as learned by the model. Indeed, a large value of this metric means that a large proportion of the neural network output sensitivity is explained by the considered variable. It can also be used to filter out the insignificant features: a very small value of this metric means that the model outcome is almost insensitive to the feature.


\subsection{Local feature importance}

In this subsection, we are describing a method for capturing the local relative importance of input features in a small neighborhood of one sample of interest. This local importance is particularly useful if one would like to understand what are the main factors behind the prediction of the neural network at a specific observation. 
By first order Taylor expansion, we know that any differentiable function can be locally approximated by a linear combination of its inputs weighted by its derivatives.
More formally, for any input vector $x$ close to $x^0$ we have:
$$f(x)- f(x^0) =  \sum_{j=1}^p \frac{\partial f(x^0)}{\partial x_j}(x_j - x^0_j) + \sum_{j=1}^p o(x_j - x_j^0)$$

where $o(x_j - x_j^0) = h(x_j)(x_j - x_j^0)$ with $h(x_j) \to 0$ when $x_j \to x_j^0$.

This previous decomposition shows that in a small neighborhood of $x^0$, the knowledge of the partial derivatives of the neural network function $f$ is enough to explain its behaviour.
The local importance of input feature $j$  at the sample $x^0$ of a non sequential type of model is then defined as follows:
$$\lambda^0_j = \frac{100}{C}\Bigg(\frac{\partial f(x^0)}{\partial x_j}\Bigg)^2$$
where $C$ is the normalization factor so that $\sum_{j=1}^p \lambda_j^0 = 100$ and with $p$ the number of input features.
Intuitively $\lambda^0_j$ represents the percentage of the model sensitivity at $x^0$ due to variable $j$. This definition can be trivially adapted to the two other types of architectures described in \ref{SS::2.2}.

\subsection{Time lag importance}

In order to take into account the time component for sequential types of networks, we propose a lag sensitivity analysis in addition to the feature importance analysis. We are interested in determining what are the most influential time lags on the current observation. To measure the influence of the $k$-th lag, we propose the following metric:
$$\gamma_{k} = \frac{100}{Kp} \sum_{j=1}^p \sqrt{\frac{1}{T}\sum_{t=1}^T \Bigg(\frac{\partial y^t}{\partial x^{t-k}_j}\Bigg)^2}$$ with $ 0 \leq k \leq \tau -1$ and $K$ the normalization factor so that $\sum_{k=0}^{\tau -1} \gamma_k = 100$. There are two levels of aggregation, one across all the $T$ sequences and one across all the $p$ features.

Intuitively, $\gamma_k$ represents the percentage of the overall time sensitivity of the model due to lag $k$. As explained in the previous part, the main use of this metric is to rank the time lags by predictive power as learned by the model. It can also be used to filter out the insignificant lags. Indeed, a very small value of this metric means that the model outcome is almost insensitive to this time lag.

This metric can also be defined locally to analyze the time sensitivity of the model at a chosen sequence $x^t$. 

\subsection{Categorical variables}

In order to be used by a neural network model, categorical features have to be transformed into numerical variables via one-hot encoding or embedding. This is because, as mentioned in \ref{SS::2.1}, neural networks are inherently differential and can only handle continuous and numerical features. This means that, internally, categorical features are viewed as continuous by the neural network and hence all the previously described metrics are well suited to assess the importance of categorical input features.

\section{Simulation}
\label{S::3}

In order to confirm that our method indeed captures an intuitive notion of feature importance, we test it on a simulated regression setting where we can control the relative importance of the inputs. 

We consider the following data generating process:
$$Y = \cos(X_1) + \sin(X_2) + 2X_3 + X_4 + \frac{1}{100}X_5 + \epsilon$$ 
The outcome $Y$ is generated from 5 input features distributed as follow: 
$$X_1, X_2, X_3, X_4, X_5 \stackrel{iid}{\sim} \mathcal{N}(0,1)$$ 
and from a regression noise $\epsilon \sim \mathcal{N}(0, 0.01)$.

This regression function has been chosen to be complex enough and especially non-linear to justify the use of a neural network model over a linear one but it remains additive in its inputs to keep the importance of each feature clear from its definition. 

We generate a dataset of 10,000 observations and use 85\% to train a fully-connected neural network with two-hidden layers of 64 and 32 units respectively and ReLU activation function by minimizing the mean-squared error (MSE). We obtain an out-of-sample MSE of $1.269 \times 10^{-2}$ which proves good convergence of the neural network. We then compute the global features importance of the fitted network as defined in \ref{SS::2.3} on the training dataset.

\begin{figure}[h]
    \centering
  \includegraphics[width=7cm]{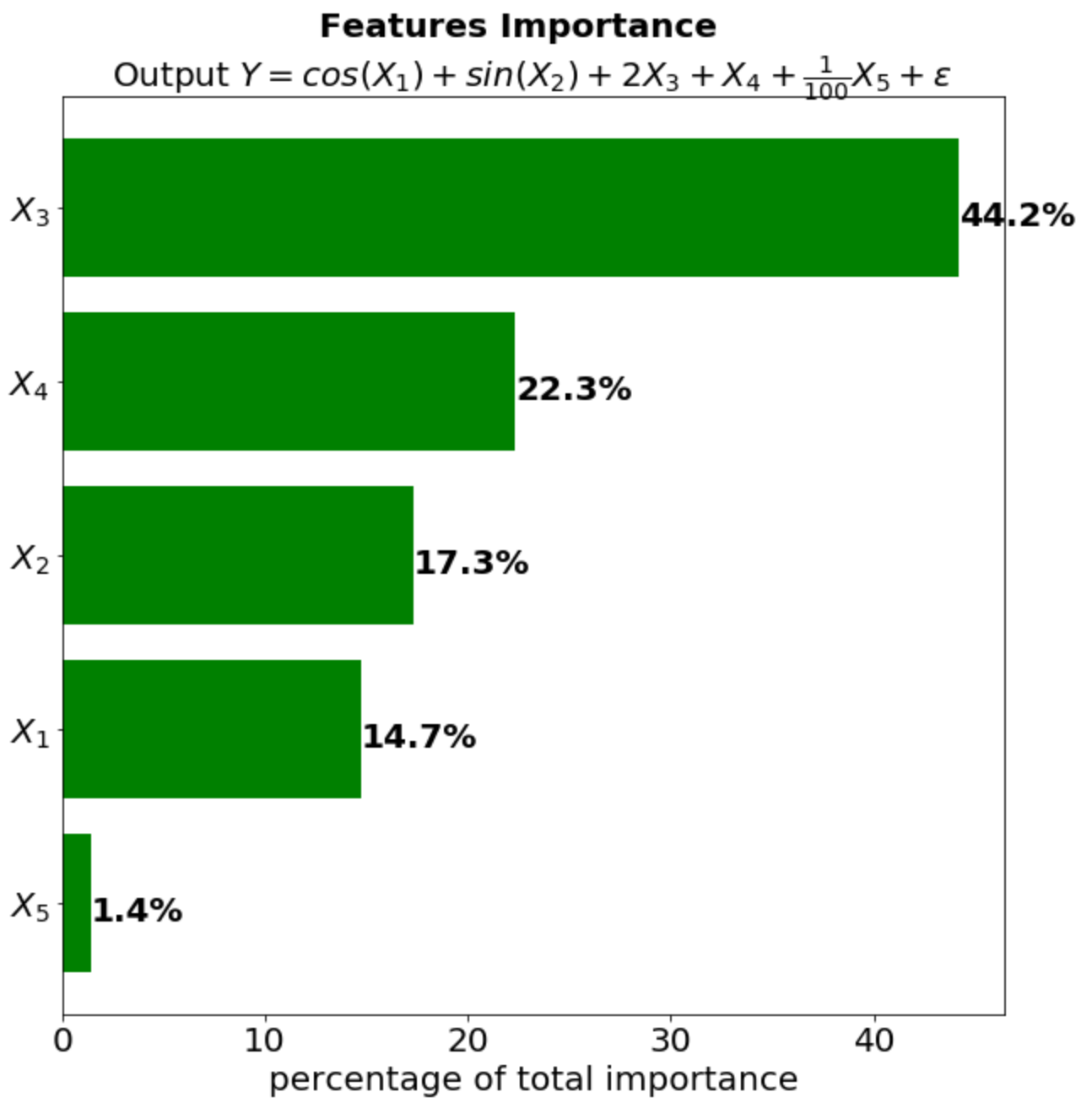}
  \caption{Ranking of global features importance of the simulated regression}
  \label{fig1}
\end{figure}

We can see on Figure \ref{fig1} that the importance of variable $X_3$ is twice as big as variable $X_4$ as expected. The impact of the terms $\cos(X_1)$ and $\sin(X_2)$ are approximately equivalent which is also intuitive. Finally, the variable $X_5$ has an insignificant effect which is consistent since its magnitude is of the same order as the regression noise. 

\section{Experiment}
\label{S::4}


\begin{figure}[!htb]
\centering
    \subfloat[Our method]{%
        \includegraphics[width=0.475\textwidth]{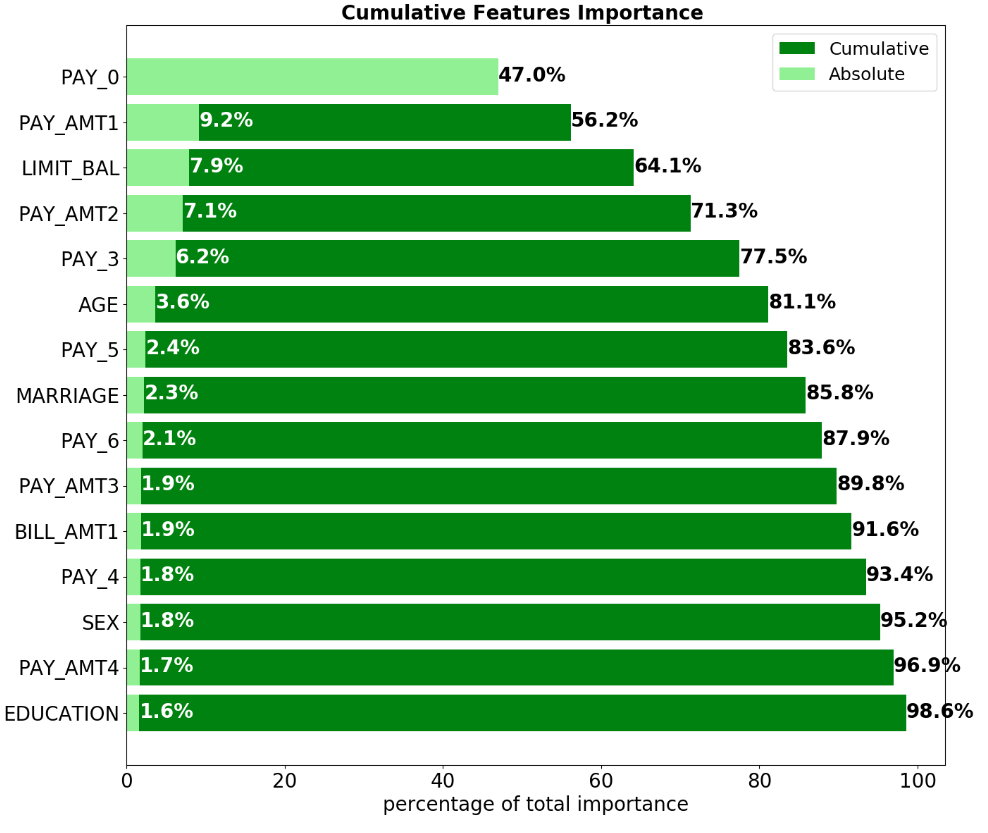}%
        \label{Our method}%
        }%
    \hfill%
    \subfloat[LIME]{%
        \includegraphics[width=0.525\textwidth]{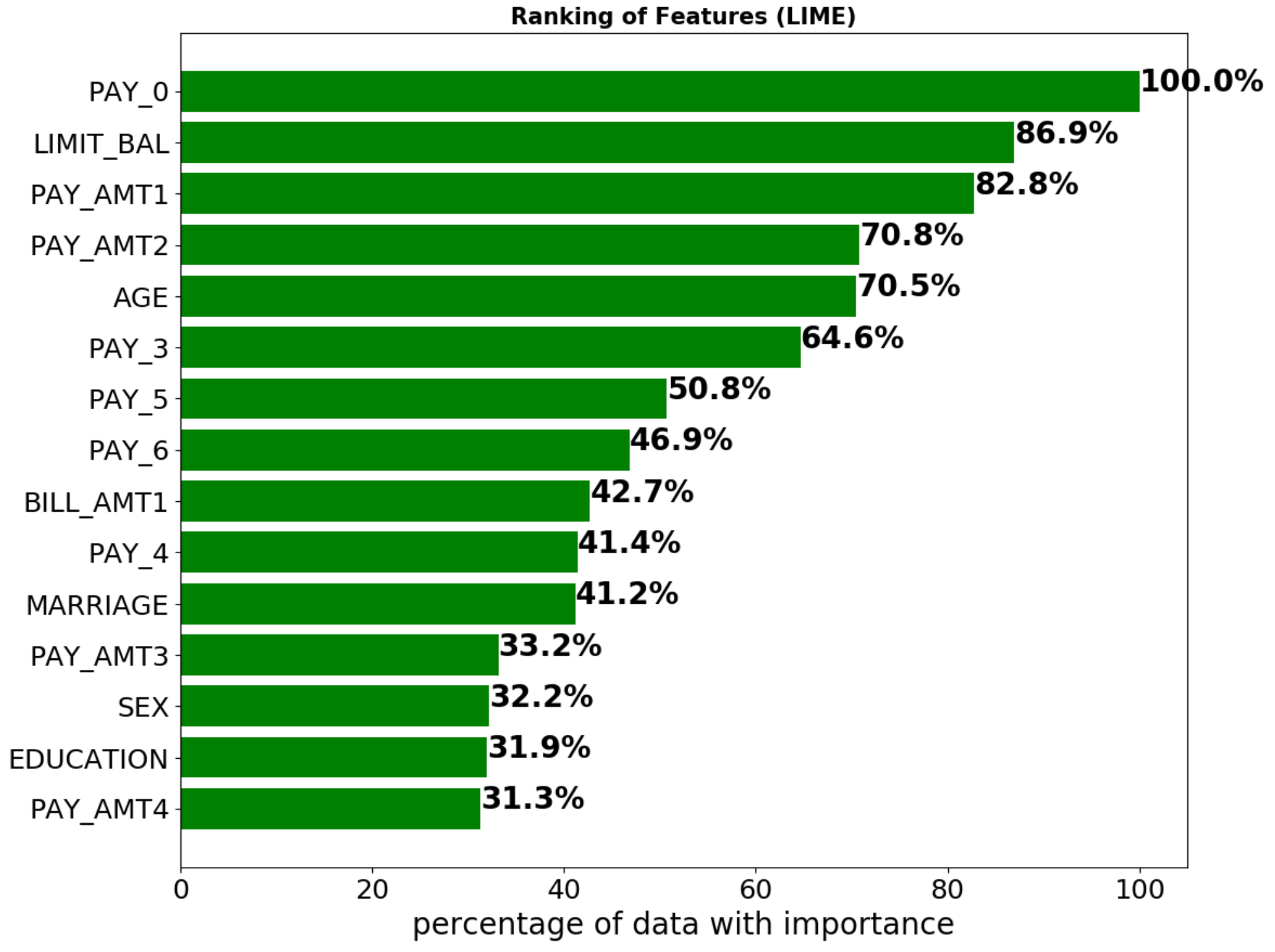}%
        \label{LIME}%
        }%
    \caption{Global features importance of FCN}
    \label{fig2}
\end{figure}

\begin{figure}[!htb]
\centering
    \subfloat[Decision Tree]{%
        \includegraphics[width=0.49\textwidth]{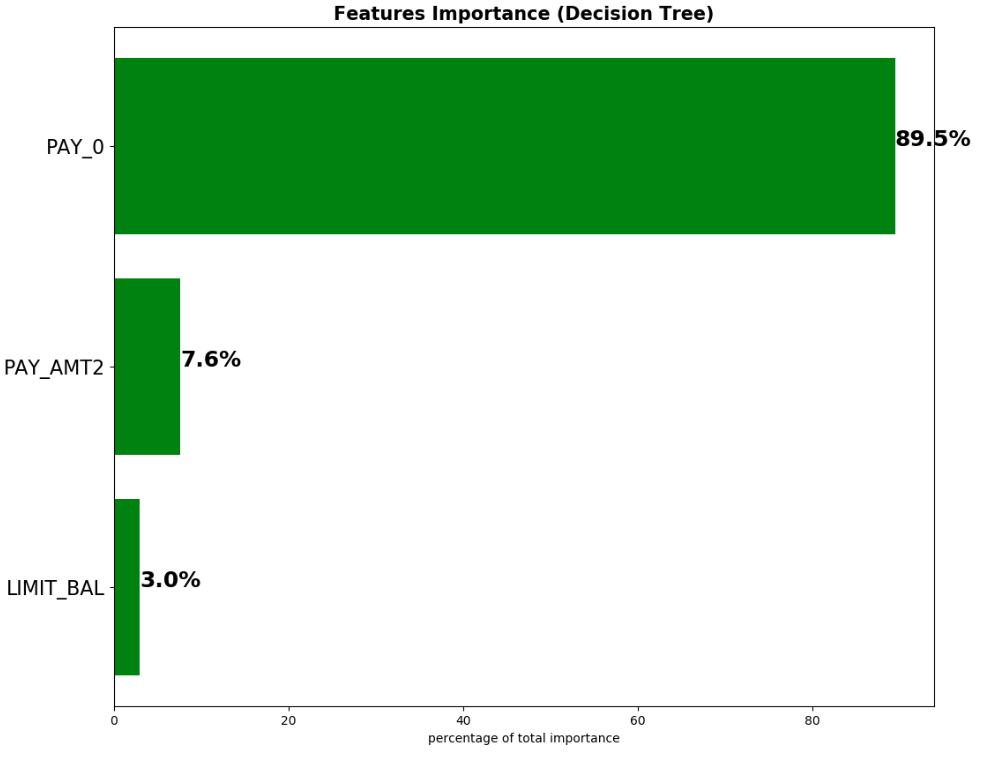}%
        \label{Decision Tree}%
        }%
    \hfill%
    \subfloat[Logistic regression]{%
        \includegraphics[width=0.5\textwidth]{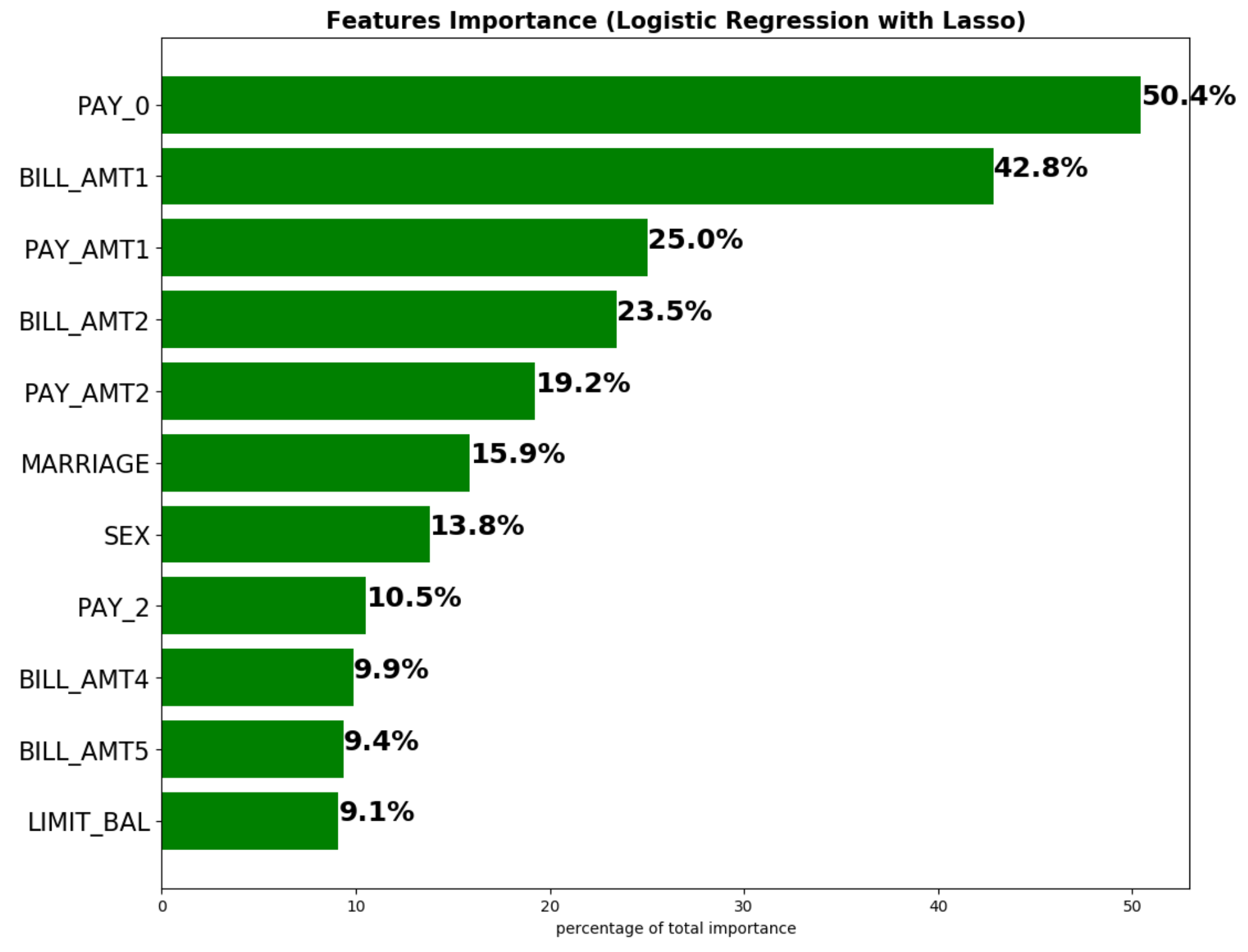}%
        \label{Logistic regression}%
        }%
    \caption{Global features importance of decision tree and logistic regression}
    \label{fig4}
\end{figure}

In this section, we evaluate the performance of the proposed neural networks interpreter on a real dataset of customers credit card defaults obtained from the UC Irvine Machine Learning Repository \cite{default}. This dataset contains 30,000 observations of credit card payments. For each observation, the binary outcome of default event is recorded along with 23 input features that describe the customer and his payment history. A description of the input features can be found in the reference of the dataset.

We train a fully-connected neural network (FCN) on this dataset and use our method to obtain the importance of the input features as learned by the network and compare it with the features importance generated by the LIME method. Additionally, we also compare the features importance of the neural network with the ones from conventional explainable machine learning models such as decision tree and logistic regression. All the models are trained on 25,000 samples and evaluated on 5,000 samples. We train the FCN model by minimizing the cross-entropy using Adam optimizer with a learning rate of 0.002 and a decay of 0.001 and using a $\ell_1$ penalty with weight 0.01. The architecture consists of one hidden layer of 64 units with hyperbolic tangent activation function and a softmax output layer of 2 units. We limit the total number of training epochs to 100 and also use early stopping on a validation set to avoid overfitting.

Figure \ref{fig2} shows the global features importance ranking of the FCN generated by our method and the LIME method. Since LIME only generates local features importance for each instance, we count the number of time each feature is selected by LIME across all the instances to compute its global features importance. As shown in Figure \ref{fig2} (a), the top 5 most important features obtained from our method are $PAY\_0$, $PAY\_AMT1$, $LIMIT\_BAL$, $PAY\_AMT2$ and $PAY\_3$. As shown in Figure \ref{fig2} (b), LIME selects $PAY\_0$, $LIMIT\_BAL$, $PAY\_AMT1$, $PAY\_AMT2$ and $AGE$ as the top 5 most important features for the same fitted FCN model. Overall, the global features importance ranking of both methods is comparable. The two approaches share similar top important features such as $PAY\_0$, $PAY\_AMT1$ and $LIMIT\_BAL$.  Moreover, our interpreter outperforms LIME in terms of time efficiency for generating global features importance for neural networks. It only takes 0.5 seconds compared to 6 hours for LIME for 25,000 samples on neural networks, which sets our interpreter apart from LIME on generating global features importance.

We further compare our method with explanations from more interpretable machine learning models. We train a decision tree and a logistic regression model on the same data set so that they achieve similar performance than the ones reported in \cite{defaultpaper}. We fit a decision tree of depth 4 and define its feature importance as the weighted impurity decrease. As shown in Figure \ref{fig4}(a), the tree picks up $PAY\_0$, $PAY\_AMT2$ and $LIMIT\_BAL$ as the most important features which is similar to the FCN interpretations. We also fit a logistic model with a Lasso regularization and define its feature importance as the magnitude of its coefficients. Figure \ref{fig4}(b) illustrates consistent top features importance ranking compared to the FCN model.

In addition of ranking the importance of input features, our interpreter can also be used in selecting significant features and hence removing redundant information. Given the cumulative features importance ranking from our interpreter in Figure \ref{fig2}(a), we can select the top features that explain $90\%$ of the overall features importance, this represents around 10 variables out of the 23. In order to compare the performance of a smaller FCN network (32 units and no regularization) fitted on the selected subset of the variables from the original one, we train 30 times on the same training set the two networks and evaluate their performances by recording their classification error rates on the same testing set. This allows to take into account the variance of performance due the randomness of the optimization procedure (random initialization of the networks' weights and random batches). As shown in Table \ref{tab1}, the FCN fitted on the subset of inputs achieves a similar testing error rate than the original network which illustrates that the variables filtered out are indeed not adding any significant predictive power.

\begin{figure}[h]
    \centering
  \includegraphics[width=7cm]{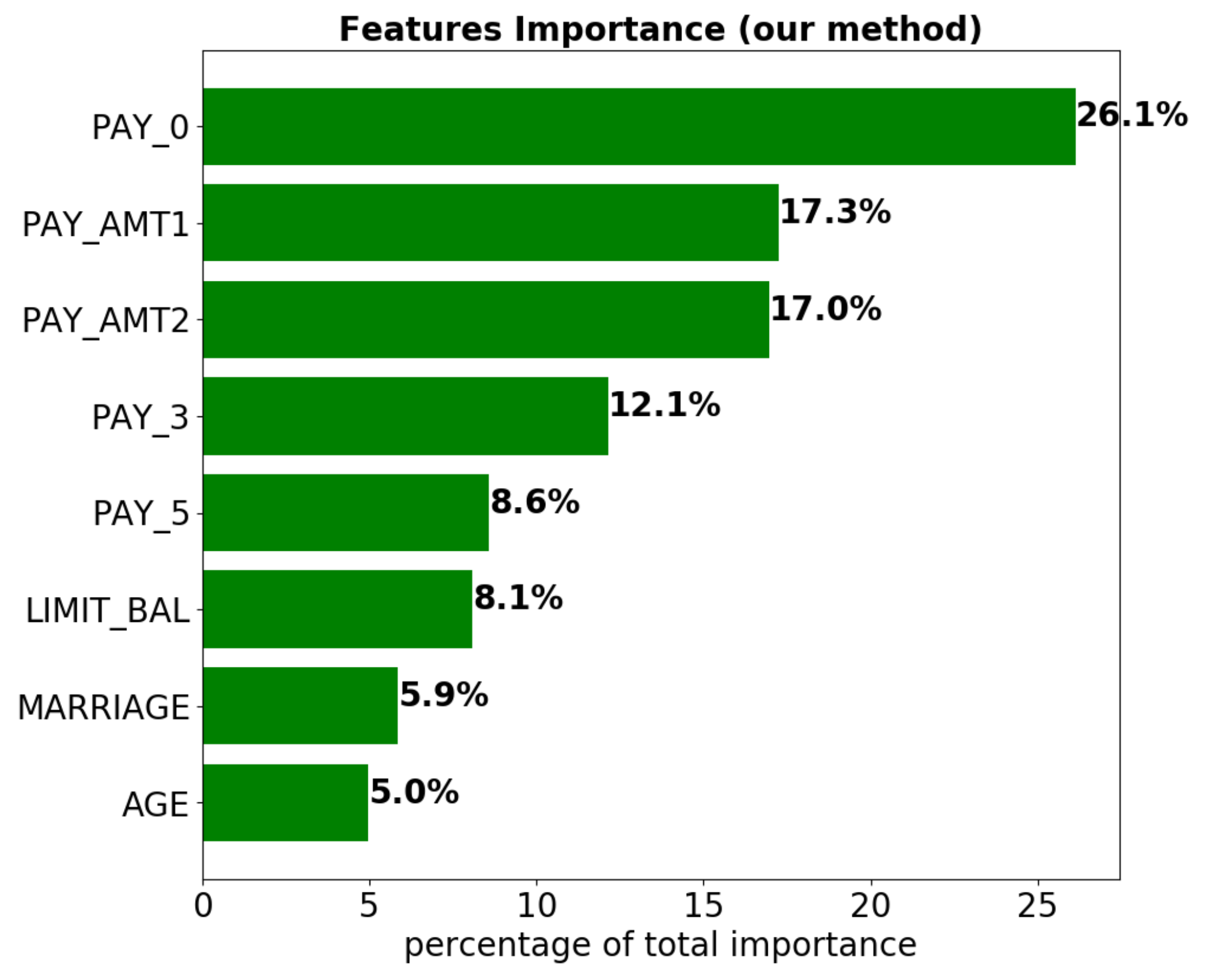}
  \caption{Ranking of global features importance after features selection (8 features)}
  \label{fig3}
\end{figure}

\begin{table}[!htb]
  \caption{Comparison of FCN performance before and after features selection over 30 runs}
  \centering
  \begin{tabular}{lll}
    Model     & Training error rate     &  Testing error rate \\
    \midrule
    FCN $\ell_1$ with all 23 features & 17.93\% $\pm$ \:0.000352   & 18.66\% $\pm$ \:  0.000484    \\
    FCN with top 90\% features importance     & \textbf{17.89\%} $\pm$  \:0.000346 & \textbf{18.47\%} $\pm$ \:0.000834      \\
    \bottomrule
  \end{tabular}
  \label{tab1}
\end{table}

\section{Conclusion and Future work}

In this paper, we propose sensitivity based neural network explanations. Using the derivatives of the model outputs with respect to its inputs, the proposed interpreter is able to generate the relative features importance in an efficient manner. We evaluate the performance of the interpreter comprehensively on both synthetic and real data. The experiment results demonstrate that our interpreter achieves intuitive feature importance ranking and similar to LIME, while being much faster. It also generates comparable explanations to intrinsically interpretable models like decision tree and logistic regression. Additionally, this interpreter can help to filter insignificant features allowing the use of simpler and faster to train models. In the future, we would like to incorporate second order derivatives in the interpreter to take into account interactions between features and also extend this approach to other "black-box" machine learning models.

\bibliography{biblio}
\bibliographystyle{plainnat}

\end{document}